\documentclass{article}
\usepackage[main,final]{neurips_2026}

\usepackage[utf8]{inputenc} 
\usepackage[T1]{fontenc}    
\usepackage{url}            
\usepackage{booktabs}       
\usepackage{amsfonts}       
\usepackage{nicefrac}       
\usepackage{microtype}      
\usepackage{xcolor}         
\usepackage{graphicx}
\usepackage{tabularx}
\usepackage{soul}
\usepackage{multirow}
\usepackage{algorithm}
\usepackage{algorithmic}
\usepackage{adjustbox}
\usepackage{colortbl}
\usepackage{amsmath}
\usepackage[hidelinks,breaklinks=true]{hyperref}
\title{FuseAdapt: Parameter-Efficient Multimodal Fusion for Semantic Segmentation with Missing Modalities}

\author{Xin Zhang \\
National University of Singapore \\
\texttt{x.zhang@u.nus.edu}
\And
Robby T. Tan \\
National University of Singapore \\
\texttt{robby.tan@nus.edu.sg}}

\hypersetup{pdftitle={FuseAdapt: Parameter-Efficient Multimodal Fusion for Semantic Segmentation with Missing Modalities},pdfauthor={Xin Zhang and Robby T. Tan}}
\makeatletter
\AtBeginDocument{\let\hyper@natlinkstart\@gobble\let\hyper@natlinkend\@empty\let\hyper@natlinkbreak\@firstoftwo}
\makeatother
\begin{document}
\maketitle

\begin{abstract}
Multi-modal semantic segmentation benefits from complementary sensors, but this benefit is fragile when some modalities are degraded or missing at test time.
Many existing methods rely on dedicated fusion modules or multi-branch cross-modal encoders, and the fusion often becomes unreliable when modalities are absent, leading to substantial performance drops.
We propose \textbf{FuseAdapt}, a parameter-efficient framework that performs fusion within the adaptation space of a frozen vision foundation model (VFM), exploring whether VFM priors can serve as an effective shared backbone for multi-modal segmentation.
Instead of attaching a standalone fusion network, FuseAdapt assigns each modality a lightweight adaptation branch and performs \emph{adaptation-space fusion} by keeping Q/V low-rank adaptation private to each modality and routing only FFN low-rank residual activations across modalities inside the frozen backbone.
The pretrained VFM remains fixed while modality-specific corrections and explicit cross-modal interaction are learned through the adaptation paths.
To handle missing modalities, we further introduce \textbf{Predictive Modality Modeling (PMM)}, a direct representation-level objective for cross-modal redundancy.
During training, PMM masks a target modality and predicts its embedding from the visible ones, supervised by a teacher target computed under full-modality input.
PMM is used only during training and adds no inference-time overhead.
Across three multi-modal segmentation benchmarks (MCubeS, DELIVER, MUSES), FuseAdapt consistently improves both modality-complete and modality-incomplete segmentation with a small trainable budget, yielding up to +3.01 mIoU under full modalities and +11.25 mIoU under missing modalities over prior state-of-the-art.
Project page: \url{https://devinxzhang.github.io/FuseAdapt-Page/}.
\end{abstract}

\section{Introduction}
\label{sec:intro}

Scene understanding often benefits from combining multiple sensing modalities~\citep{huang2022multi,jia2023event,park2017rdfnet,zhang2018sleep}.
RGB~\citep{wei2024stronger,zhang2025mamba,cyclegrpo2026,zhang2024omgllava,li2025denseworld,yuan2026sa2va,zhou2026samtok,zhang2024heap,zhang2023adaptive}, depth~\citep{cao2021shapeconv,song2022improving,ying2022uctnet}, LiDAR~\citep{yan20222dpass,wang2022multimodal,li2023mseg3d}, and event streams~\citep{xia2023cmda,zhang2021issafe,li2025brain} provide complementary cues about appearance, geometry, and motion.
In practice, however, this multi-sensor benefit is fragile: a modality may be noisy~\citep{yang2025restore,yang2026unipet}, degraded, spatially imperfect, or unavailable at inference time.
Such cases are common in autonomous driving~\citep{caesar2020nuscenes,li2026frequency} and embodied settings~\citep{wang2024embodiedscan}, where illumination, weather~\citep{du2026weatherreasonseg,teng2025raindropgs,yang2025erf}, texture, sensor failure, or calibration conditions~\citep{yuan2025tailored} can affect sensors unequally.
A practical multi-modal segmentation model should therefore work in two settings: modality-complete semantic segmentation (MCSS), where all modalities are available and should be fused effectively, and modality-incomplete semantic segmentation (MISS), where only a subset is available and the model should remain reliable.

\begin{figure}[t]
	\centering
	\includegraphics[width=0.9\linewidth]{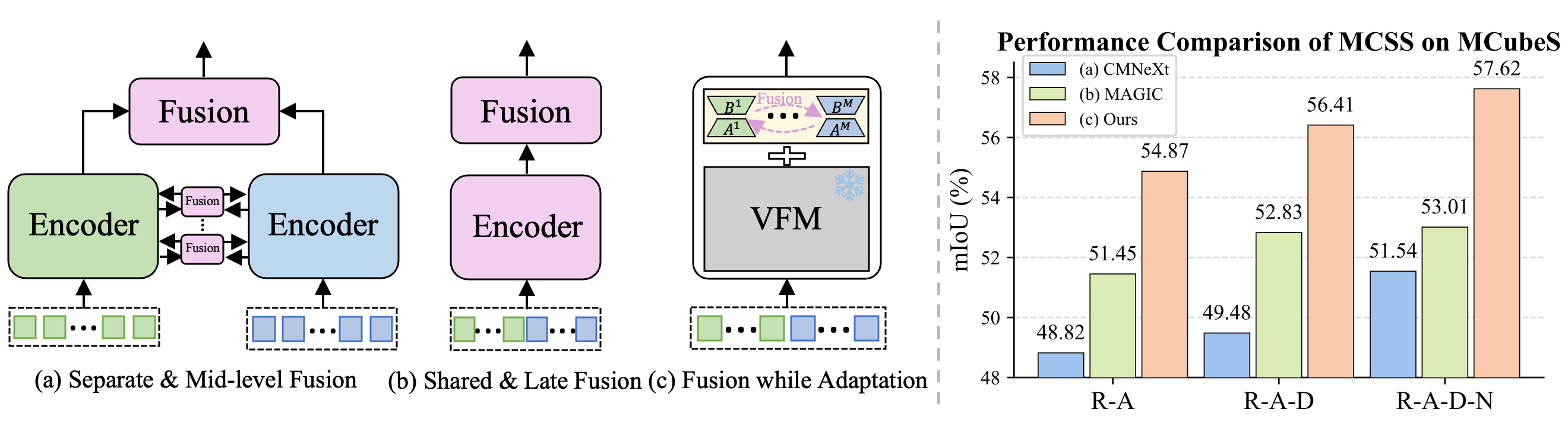}
    \caption{
Comparison of multi-modal fusion strategies.
(a) Separate encoders with mid-level fusion can replicate backbone computation across modalities.
(b) Shared-encoder or late-fusion designs reduce this cost, but can make fusion depend strongly on the dominant or cleanest modality.
(c) FuseAdapt keeps the VFM frozen and performs cross-modal exchange through trainable adaptation paths.
Right: MCSS on MCubeS, where FuseAdapt improves mIoU across modality combinations.
}
    \label{fig:mss}
\end{figure}

Existing multi-modal segmentation methods often trade off modality-specific modeling, compactness, and robustness.
Separate-encoder designs fuse intermediate features from different modalities~\citep{broedermann2023hrfuser,zhang2023cmx}, but can increase parameters and computation as the number of sensors grows.
RGB-centric methods treat RGB as the main input and use other modalities as auxiliary cues~\citep{zhang2023delivering}, which can be limiting when RGB is degraded or absent.
Shared-encoder and late-fusion methods improve compactness~\citep{zheng2024centering,zheng2024learning}, but their shared stream can favor the most informative or easiest modality, making the contribution of other sensors less stable.
For missing-modality robustness, recent methods go beyond modality dropout by using text anchors~\citep{zheng2024learning}, subset consistency~\citep{zheng2024centering,zheng2024magic++}, or modality regularization~\citep{zheng2025reducing}.
These strategies provide indirect supervision for dense prediction: text anchors are coarse, and subset consistency can be affected by cross-modal mismatch.%

We study how to separate sensor-specific adaptation from cross-sensor communication within a shared frozen backbone. \textbf{FuseAdapt} keeps modality-specific Q/V low-rank residuals private and restricts receiver-conditioned, token-wise exchange to FFN low-rank residual activations (Fig.~\ref{fig:mss}(c)). The frozen dense FFN transformation stays on the receiving stream, while routing is normalized over the available sensors. We call this constraint on the trainable communication path \emph{adaptation-space fusion}. The exchanged tensors remain intermediate features; the contribution is the separation of private attention adaptation and cross-modal residual communication, rather than a new tensor domain or the individual use of LoRA and routing. A shared VFM supplies the pretrained representation, and controlled experiments with MiT backbones test whether the benefit depends on that representation.

\begin{figure}[t]
	\centering
	\includegraphics[width=0.9\linewidth]{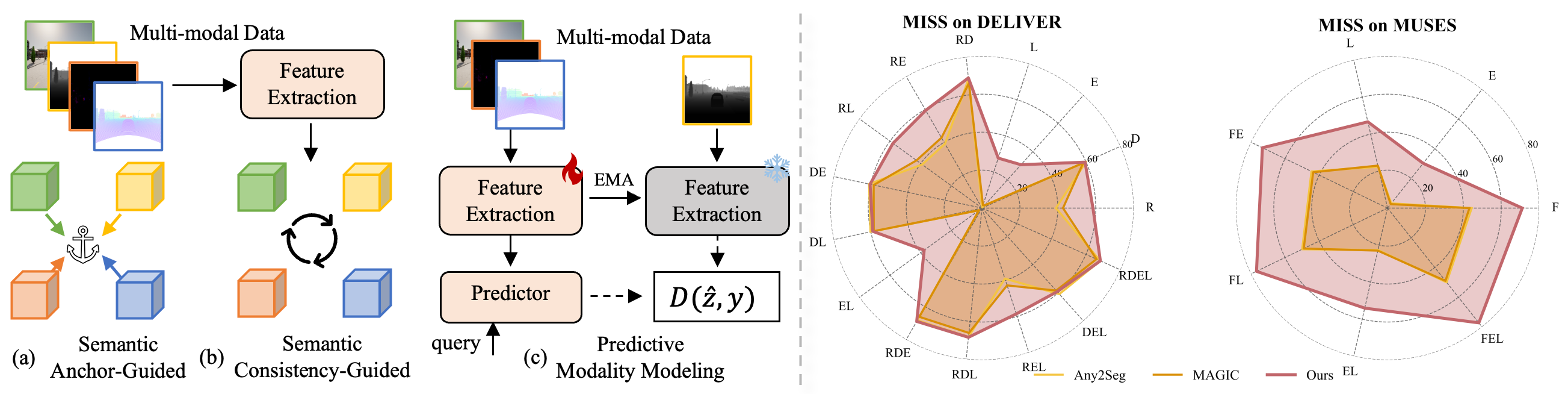}
    \vspace{-2mm}
    \caption{
    Comparison of designs for robustness under missing modalities (MISS).
    (a) Semantic-anchor methods align modality features to text embeddings, but text anchors can be coarse for pixel-level segmentation.
    (b) Semantic-consistency methods enforce agreement across different modality subsets, but such constraints can be affected by cross-modal mismatch.
    (c) Predictive Modality Modeling (PMM) casts MISS training as conditional representation prediction. Learnable query tokens aggregate cues from visible modalities to predict a missing-modality embedding, improving MISS performance on DELIVER and MUSES.
    }
    \label{fig:miss}
\end{figure}

To handle missing modalities, we introduce \textbf{Predictive Modality Modeling (PMM)} (Fig.~\ref{fig:miss}(c)).
PMM masks a target modality during training and predicts its representation from the visible modalities, supervised by a teacher target computed under full-modality input.
PMM applies the joint-embedding predictive principle at modality level: target-specific queries predict an unobserved sensor's latent representation from visible context. Its EMA teacher supplies a temporally smoothed, stop-gradient target. After projection onto a common token grid, each query can attend to all visible positions rather than imposing same-index visible-to-target regression. This does not remove the common-grid requirement of routing and averaging. The teacher and predictor are used only during training.

We evaluate FuseAdapt on MCubeS~\citep{liang2022multimodal}, DELIVER~\citep{zhang2023delivering}, and MUSES~\citep{brodermann2024muses} under both MCSS and MISS protocols.
FuseAdapt yields up to +3.01 mIoU under full modalities and +11.25 mIoU under missing modalities over prior state-of-the-art, with ablations isolating the contributions of modality-specific adaptation, routed cross-modal exchange, PMM, and the choice of frozen VFM.
Our contributions are:
\begin{itemize}
\item FuseAdapt separates modality-private Q/V adaptation from receiver-conditioned cross-modal exchange restricted to FFN low-rank residual activations, with routing adjusted to the visible sensor subset.
\item PMM trains missing-modality robustness through target-conditioned latent prediction. Controls distinguish its prediction target from output distillation and generic fused-feature matching.
\item Experiments on three benchmarks, same-backbone PEFT and information-flow controls support the design. We additionally characterize its inference cost and the scope of its robustness.
\end{itemize}

\section{Related Work}
\label{sec:related_work}

\textbf{Multi-Modal Semantic Segmentation }
Multi-modal semantic segmentation uses complementary RGB~\citep{wei2024stronger,zhang2025mamba,qian2024msnsegnet,zhang2022gigapixel,deng2020deep}, depth~\citep{wang2020deep,ying2022uctnet,song2022improving,cao2021shapeconv,park2017rdfnet}, LiDAR~\citep{li2022deepfusion,li2023mseg3d,zhuang2021perception,yan20222dpass,wang2022multimodal}, thermal~\citep{hui2023bridging,zhang2023efficient,zhang2021abmdrnet,wu2022complementarity}, and event cues~\citep{xia2023cmda,zhang2021issafe,jia2023event}.
Prior work often uses fixed modality pairs such as RGB-depth~\citep{ying2022uctnet,song2022improving}, RGB-centric fusion~\citep{zhang2023delivering}, or shared-backbone fusion with auxiliary alignment~\citep{zheng2024learning,zheng2024centering,zheng2024magic++}.
In contrast, FuseAdapt keeps a frozen VFM and performs cross-modal exchange through trainable adaptation paths, avoiding a standalone fusion network.

\textbf{Missing or Degraded Modalities }
Existing approaches use modality dropout, cross-modal distillation, and consistency regularization~\citep{liu2024fourier,wang2023learnable,zheng2024learning,zheng2024centering,zheng2024magic++}. CRM~\citep{shin2024crm} combines complementary RGB-thermal masking with clean-to-masked self-distillation; M3L~\citep{maheshwari2024m3l} uses a full-modality teacher to supervise masked-modality segmentation. PMM instead conditions on a designated missing sensor and predicts its full-context latent representation. This follows the latent-context prediction principle of I-JEPA~\citep{assran2023ijepa}, applied to modalities rather than image regions. Masking, EMA, and teacher--student optimization are established mechanisms; the prediction target and modality conditioning distinguish our objective.

\textbf{Parameter-Efficient Adaptation and Fusion }
Prompt tuning, adapters, and LoRA reduce the cost of adapting pretrained models~\citep{lester2021power,jia2022visual,houlsby2019parameter,chen2022adaptformer,hu2022lora,yan2025erlora}. StitchFusion~\citep{li2024stitchfusion} transfers dense features between encoder streams through MultiAdapters. MLE-SAM~\citep{zhu2024customize} combines modality-specific LoRA experts and routing with multi-scale feature/head fusion. MoLE~\citep{wu2024mole} composes task/concept LoRAs within a single stream. Concurrent M4-SAM~\citep{liu2026m4sam} combines modality-dispatched convolutional experts with gated feature fusion for RGB-D video, while MILE~\citep{muralidhara2026mile} selects isolated continual-task experts. FuseAdapt instead keeps Q/V residuals private and restricts directed source-to-receiver communication to FFN low-rank residuals among currently visible sensors. Section~\ref{sec:controlled} tests this constraint against routing all low-rank paths or complete FFN features.

\begin{figure*}[t]
	\centering
	\includegraphics[width=0.9\linewidth]{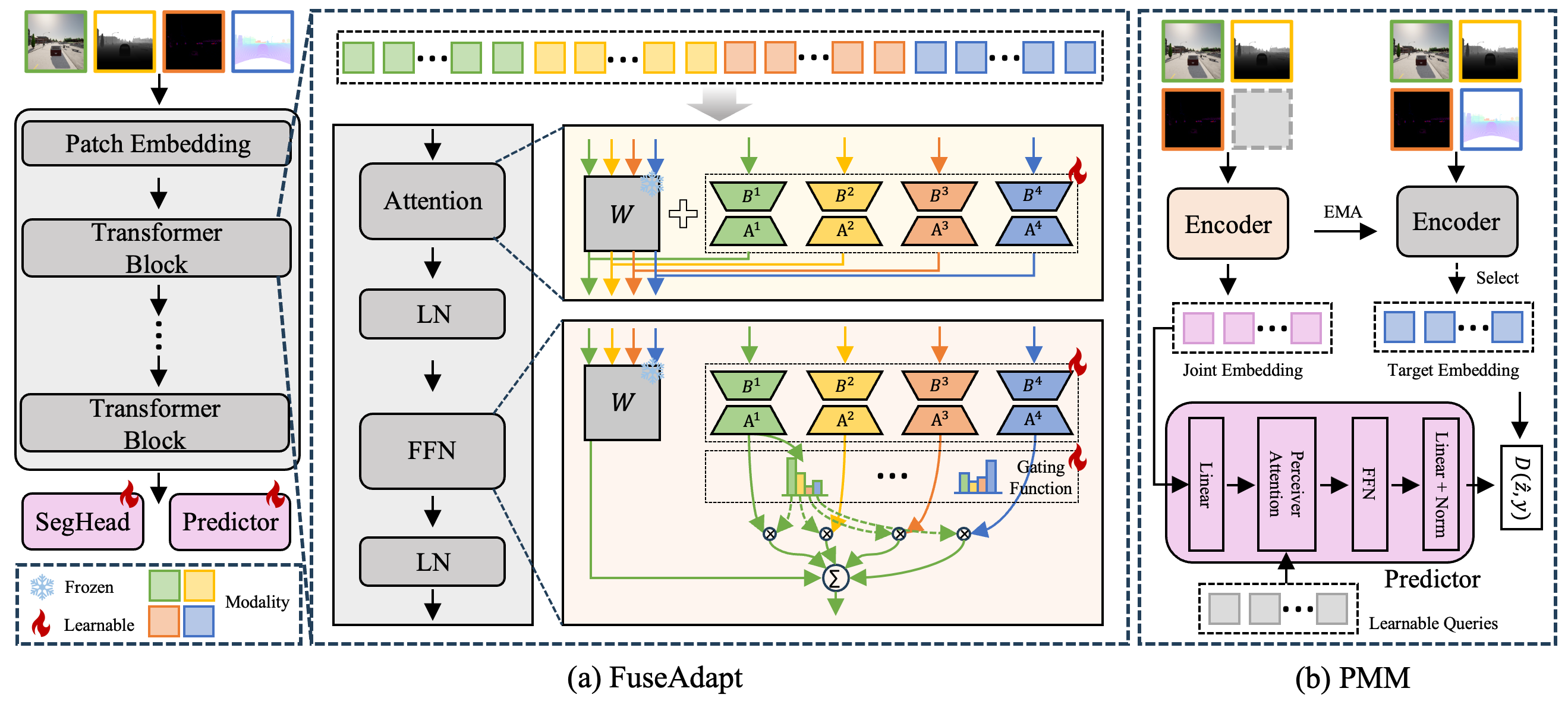}
    \vspace{-2mm}
\caption{
\textbf{Overview of our FuseAdapt framework.}
(a) \emph{Adaptation-Space Fusion:} Multi-modal inputs (e.g., RGB, depth, event, LiDAR) share a frozen VFM backbone. Each modality attaches a lightweight adaptation branch inside the transformer blocks, and a receiver-conditioned router mixes only FFN low-rank residual activations. Fusion is thus performed inside the adaptation path while keeping the backbone frozen, and the fused features are sent to a shared segmentation head. For clarity, we show the routing for one modality; the others follow the same design.
(b) \emph{PMM:} To handle missing modalities, an EMA teacher provides the target embedding for a selected modality. A student and a Perceiver-style predictor use the visible modalities to predict this embedding via learnable queries. PMM is trained with a distance loss $D(\hat{z}, z)$.
}
	\label{fig:main_fig}
\end{figure*}

\section{Proposed Method}
\label{sec:method}

We address robust multi-modal semantic segmentation in two regimes: modality-complete semantic segmentation (MCSS), where all modalities are available, and modality-incomplete semantic segmentation (MISS), where only a subset is available at test time. Our goal is to adapt a frozen vision foundation model (VFM) into a multi-modal learner that can fuse available modalities with a small trainable budget and learn redundancy for missing-modality robustness.

As shown in Fig.~\ref{fig:main_fig}, the input contains four aligned image-like modalities of the same scene: RGB $\mathbf{R}\in\mathbb{R}^{h\times w\times 3}$, depth $\mathbf{D}\in\mathbb{R}^{h\times w\times C^D}$, LiDAR projection $\mathbf{L}\in\mathbb{R}^{h\times w\times C^L}$, and event stack $\mathbf{E}\in\mathbb{R}^{h\times w\times C^E}$, with pixel labels $\mathbf{Y}$ over $K$ classes. Following~\citep{zhang2023delivering}, we set $C^D=C^L=C^E=3$ and feed all modalities as image-like inputs to a shared frozen VFM.

FuseAdapt has two components. First, an \emph{adaptation-space fusion design} (\S\ref{sec:adapt}) performs fusion in the adaptation space of the frozen VFM. Each modality has lightweight trainable adaptation paths, and cross-modal exchange is implemented by routing FFN low-rank residual activations across modalities inside Transformer blocks. The pretrained backbone remains fixed. Second, \emph{Predictive Modality Modeling (PMM)} improves MISS by training the model to predict a masked modality representation from visible modalities, using a teacher target computed with full-modality context. PMM is used only during training and discarded at inference.

\subsection{Adaptation-Space Fusion}
\label{sec:adapt}

We realize adaptation-space fusion through modality-specific low-rank adaptation paths inside the frozen VFM, with cross-modal exchange implemented by routed FFN low-rank residual activations. We keep self-attention adaptation modality-specific to preserve sensor-specific correction, and place cross-modal exchange in the FFN adaptation path.

\noindent\textbf{Self-Attention Adaptation }
Let $\mathcal{M}=\{r,d,e,l\}$ denote the modality set. At a Transformer block, let $\mathbf{X}^m\in\mathbb{R}^{N\times d}$ be the token features of modality $m$, where all modalities have the same projected token count $N$ at this block and $d$ is the hidden dimension. Routing and averaging require this common grid; arbitrary native token lengths are not supported. We omit the block index and bias terms for clarity. For each modality, we add trainable low-rank updates only to the query and value projections:
\begin{equation}
\mathbf{Q}^m=\mathbf{X}^m(\mathbf{W}_Q+\Delta\mathbf{W}_Q^m),\quad
\mathbf{V}^m=\mathbf{X}^m(\mathbf{W}_V+\Delta\mathbf{W}_V^m),
\end{equation}
where $\mathbf{W}_Q,\mathbf{W}_V\in\mathbb{R}^{d\times d}$ are frozen pretrained weights. The updates are parameterized as
\begin{equation}
\Delta\mathbf{W}_Q^m=\mathbf{A}_Q^m\mathbf{B}_Q^m,\quad
\Delta\mathbf{W}_V^m=\mathbf{A}_V^m\mathbf{B}_V^m,
\label{lora}
\end{equation}
where $\mathbf{A}_Q^m,\mathbf{A}_V^m\in\mathbb{R}^{d\times r}$, $\mathbf{B}_Q^m,\mathbf{B}_V^m\in\mathbb{R}^{r\times d}$, and $r\ll d$. The key and output projections, $\mathbf{W}_K$ and $\mathbf{W}_O$, remain frozen and shared. The attention output is
\begin{equation}
\text{MHSA}^m(\mathbf{X}^m)
=
\text{softmax}\left(
\frac{\mathbf{Q}^m(\mathbf{X}^m\mathbf{W}_K)^\top}{\sqrt{d}}
\right)
\mathbf{V}^m\mathbf{W}_O .
\end{equation}
This keeps attention adaptation separated by modality while preserving the pretrained attention structure.

\noindent\textbf{Feedforward Routing Fusion }
Cross-modal exchange is introduced in the FFN adaptation path. Let $\mathbf{W}_1\in\mathbb{R}^{d\times l}$ and $\mathbf{W}_2\in\mathbb{R}^{l\times d}$ be the frozen FFN weights. Each modality has trainable low-rank FFN updates:
\begin{equation}
\Delta\mathbf{W}_1^m=\mathbf{A}_1^m\mathbf{B}_1^m,\quad
\Delta\mathbf{W}_2^m=\mathbf{A}_2^m\mathbf{B}_2^m,
\end{equation}
where $\mathbf{A}_1^m\in\mathbb{R}^{d\times r}$, $\mathbf{B}_1^m\in\mathbb{R}^{r\times l}$,
$\mathbf{A}_2^m\in\mathbb{R}^{l\times r}$, $\mathbf{B}_2^m\in\mathbb{R}^{r\times d}$, and $r\ll\min(d,l)$.

Let $\mathcal{V}\subseteq\mathcal{M}$ be the nonempty available subset. For receiving modality $m\in\mathcal{V}$ and token $i$, router logits $g^m(\mathbf{X}^m)_i$ define
\begin{equation}
\alpha^m_{ij}=\frac{\exp(g^m(\mathbf{X}^m)_{ij})}{\sum_{k\in\mathcal{V}}\exp(g^m(\mathbf{X}^m)_{ik})},\quad j\in\mathcal{V},
\qquad \alpha^m_{ij}=0\quad(j\notin\mathcal{V}).
\label{eq:alpha}
\end{equation}
Thus unavailable sources are masked before normalization. Omitting block indices, biases, and dropout, the FFN becomes
\begin{align}
\mathbf{H}^m&=\sigma\!\left(\mathbf{X}^m\mathbf{W}_1+
\sum_{j\in\mathcal{V}}\boldsymbol{\alpha}^m_j\odot(\mathbf{X}^j\Delta\mathbf{W}_1^j)\right),\\
\mathbf{F}^m&=\mathbf{H}^m\mathbf{W}_2+
\sum_{j\in\mathcal{V}}\boldsymbol{\alpha}^m_j\odot(\mathbf{H}^j\Delta\mathbf{W}_2^j),
\end{align}
where $\sigma$ is the FFN activation and $\odot$ broadcasts token weights over channels. The routed objects are residual activations, not parameter matrices or complete FFN features. Dense transformations operate on the receiving stream; their weights remain frozen even though stream activations evolve through successive blocks. After the final block, we symmetrically aggregate visible outputs,
\begin{equation}
\hat{\mathbf{X}}_{\mathcal{V}}=\frac{1}{|\mathcal{V}|}\sum_{m\in\mathcal{V}}\hat{\mathbf{X}}^m,
\end{equation}
and pass them to the shared segmentation decoder. The same communication rule applies to full and partial inputs.

\subsection{Predictive Modality Modeling}
\label{sec:pmm}
While adaptation-space fusion handles full-modality inputs, MISS also requires the model to behave reliably when some modalities are absent. PMM addresses this by adding a representation-level prediction objective during training. Given visible modalities, the model predicts the representation of a masked target modality, supervised by a teacher target computed with full-modality context. This encourages cross-modal redundancy beyond segmentation supervision alone.

After the full-input warm-up, training mixes full and masked inputs. In the masked branch, modalities are independently retained with probability $0.5$, conditioned on a nonempty visible subset $\mathcal{V}$. When $\mathcal{V}\ne\mathcal{M}$, we sample a target $t\in\mathcal{M}\setminus\mathcal{V}$ uniformly and aggregate visible features $\hat{\mathbf{X}}^m\in\mathbb{R}^{N\times C}$ as above. When all modalities are visible, segmentation supervision remains active and no missing-target loss is computed. Appendix~\ref{app:pmm_details} gives the masking procedure.

The teacher target $y_t\in\mathbb{R}^{N\times C}$ for the masked modality $t$ is produced by an EMA teacher that takes the \emph{full-modality} input. The teacher shares the frozen VFM with the student (no duplication of backbone weights) and maintains an EMA copy only of the adaptation branches. Letting $\theta_s$ denote the student FuseAdapt parameters and $\theta_\tau$ those of the teacher, the EMA update is
\begin{equation}
\theta_\tau \leftarrow \mu\,\theta_\tau + (1-\mu)\,\theta_s,\qquad \mu=0.999,
\end{equation}
applied at every iteration. The target is stop-gradient. Full context is a useful training prior on average, not a guarantee of superior predictions for every sample, especially when a present sensor is corrupted (Appendix~\ref{app:teacher}).

\noindent\textbf{Query-Based Prediction }
A per-token linear regressor imposes same-index visible-to-target correspondence on the common grid. We instead use modality-specific learnable queries $Q_t\in\mathbb{R}^{N\times C}$, each of which attends globally to $\hat{\mathbf{X}}_{\mathcal{V}}$ through a Perceiver-style predictor:
\begin{equation}
\hat{z}_t=g_\phi(Q_t,\hat{\mathbf{X}}_{\mathcal{V}}),\qquad
\hat{z}_t\in\mathbb{R}^{N\times C}.
\end{equation}
This permits retrieval from any visible position while retaining a common target grid. It does not natively align different fields of view or temporal offsets.
The PMM loss is a token-wise cosine distance:
\begin{equation}
\mathcal{L}_\text{pmm}
=
1-\frac{1}{N}\sum_{i=1}^{N}
D\big(\hat{z}_t^{(i)},\operatorname{sg}[y_t^{(i)}]\big),
\end{equation}
where $D(\cdot,\cdot)$ denotes cosine similarity.
The full training objective is
\begin{equation}
\mathcal{L}=\mathcal{L}_\text{seg}+\mathcal{L}_\text{pmm}.
\end{equation}
At inference time, PMM is removed and the model uses the available modalities directly.

\begin{table}[!t]
    \centering
    \caption{Results in MCSS setting on MCubeS. The best results are highlighted in \textbf{bold}. (R: RGB; A: AoLP; D: DoLP; N: NIR.)}
    \label{tab:mss_mcubes}
    \renewcommand{\tabcolsep}{1pt}
    \renewcommand{\arraystretch}{0.9}
    \small
    \begin{adjustbox}{width=0.6\columnwidth}
    \begin{tabular}{l|c|c|ccc|c}
        \toprule
        \multirow{2}{*}{Method} & \multirow{2}{*}{Backbone} & \multicolumn{1}{c|}{$\#$Trainable} & \multicolumn{3}{c|}{Modals} & \multirow{2}{*}{Mean} \\
        \cmidrule{4-6}
        & & Params (M) & RA & RAD & RADN & \\
        \midrule
        MCubeSNet~\citep{liang2022multimodal}      & MiT-B4     & 65.65 & 39.10 & 42.00 & 42.86 & 41.32 \\
        CWSAM~\citep{pu2025classwise}             & ViT-B      & 15.29 & 49.78 & 48.27 & 50.59 & 49.55 \\
        MLE-SAM~\citep{zhu2024customize}          & Hiera-B+   & 20.79 & 50.61 & 50.89 & 51.02 & 50.84 \\
        CMNeXt~\citep{zhang2023delivering}        & MiT-B2     & 58.73 & 48.82 & 49.48 & 51.54 & 49.95 \\
        Any2Seg~\citep{zheng2024learning}         & MiT-B2     & 24.73 & 48.95 & 50.67 & 51.91 & 50.51 \\
        MemorySAM~\citep{liao2025memorysam}       & Hiera-B+   & 14.50 & 51.20 & 52.20 & 52.88 & 52.09 \\
        MAGIC~\citep{zheng2024centering}          & MiT-B2     & 24.73 & 51.45 & 52.83 & 53.01 & 52.43 \\
        StitchFusion~\citep{li2024stitchfusion}   & MiT-B4     & 68.02 & 52.68 & 53.26 & 53.92 & 53.29 \\
        \midrule
        Ours (wo FuseAdapt)        & ViT-B &  9.57  & 49.15 & 49.98 & 51.16 & 50.10 \\
        Ours (shared FuseAdapt)    & ViT-B &  10.64  & 54.87 & 54.85 & 54.69 & 54.80 \\
        Ours (separate FuseAdapt)  & ViT-B & 13.86  & 54.87 & 55.26 & 56.09 & 55.41 \\
        \rowcolor{gray!20}
        Ours                       & ViT-B & 14.01  & \textbf{54.87} & \textbf{56.41} & \textbf{57.62} & \textbf{56.30} \\
        \bottomrule
    \end{tabular}
    \end{adjustbox}
\end{table}

\section{Experiments}
\label{sec:exp}

\noindent\textbf{Datasets }
We evaluate on three multi-modal segmentation benchmarks covering both modality-complete semantic segmentation (MCSS) and modality-incomplete semantic segmentation (MISS). DELIVER~\citep{zhang2023delivering} contains synchronized RGB, depth, LiDAR, and event inputs for 25-class urban-scene segmentation, with 3,983/2,005/1,897 train/val/test samples and diverse adverse conditions such as fog, rain, night, blur, exposure changes, and LiDAR jitter~\citep{yuan2026drive}. MCubeS~\citep{liang2022multimodal} contains 20-class material segmentation with RGB, NIR, DoLP, and AoLP modalities~\citep{lin2025rgb}, using 302/96/102 train/val/test samples. MUSES~\citep{brodermann2024muses} provides synchronized frame-camera, event-camera, and MEMS-LiDAR data with 2D panoptic annotations, enabling evaluation with heterogeneous sensors.

\noindent\textbf{Implementation Details }
We use AdamW and train on A5000 GPUs: 8 GPUs for DELIVER and 4 GPUs for MCubeS/MUSES. Batch size is 1 per GPU for DELIVER and 4 for MCubeS/MUSES. The initial learning rate is $4\times10^{-4}$ for DELIVER/MUSES and $3.5\times10^{-4}$ for MCubeS, with a 10-epoch warm-up at 0.1$\times$. We train for 200 epochs on DELIVER/MUSES and 500 epochs on MCubeS. Input resolutions are $1024\times1024$ for DELIVER/MUSES and $512\times512$ for MCubeS. We use standard augmentations, including random resizing (0.5--2.0), horizontal flipping, color jitter, blur, and random cropping. FuseAdapt uses rank 8 for DELIVER and MCubeS, and 4 for MUSES. We use DINOv2-B for DELIVER/MCubeS and DINOv2-S for MUSES~\citep{oquab2023dinov2} as frozen VFM backbones.

\begin{table}[!t]
    \centering
    \caption{Results in MCSS setting on MUSES. The best results are highlighted in \textbf{bold}. (F: frame camera, E: event camera, L: LiDAR.)}
    \label{tab:mss_muses}
    \small
    \renewcommand{\tabcolsep}{1pt}
    \renewcommand{\arraystretch}{1}
    \begin{adjustbox}{max width=0.6\columnwidth}
    \begin{tabular}{l|c|c|cccc|c}
        \toprule
        \multirow{2}{*}{Method} & \multirow{2}{*}{Backbone} & \multicolumn{1}{c|}{$\#$Trainable} & \multicolumn{4}{c|}{Modals} & \multirow{2}{*}{Mean} \\
        \cmidrule{4-7}
        & & Params (M) & F & FE & FL & FEL & \\
        \midrule
        CWSAM~\citep{pu2025classwise}            & ViT-B    & 15.29 & 55.41 & 41.77 & 40.69 & 49.98 & 46.96 \\
        CMNeXt~\citep{zhang2023delivering}       & MiT-B0   & 10.30 & 43.37 & 43.39 & 47.03 & 46.66 & 45.11 \\
        MLE-SAM~\citep{zhu2024customize}         & Hiera-B+ & 20.79 & 73.95 & 74.73 & 75.42 & 74.80 & 74.73 \\
        \midrule
        Ours (wo FuseAdapt)        & ViT-S & 2.42 & 62.46 & 63.00 & 64.72 & 65.14 & 63.83 \\
        Ours (shared FuseAdapt)    & ViT-S & 2.69 & 74.99 & 74.96 & 76.23 & 76.59 & 75.69 \\
        Ours (separate FuseAdapt)  & ViT-S & 3.22 & 74.99 & 74.81 & 76.13 & 76.38 & 75.58 \\
        \rowcolor{gray!20}
        Ours                       & ViT-S & 3.29 & \textbf{74.99} & \textbf{75.36} & \textbf{76.94} & \textbf{77.28} & \textbf{76.14} \\
        \bottomrule
    \end{tabular}
    \end{adjustbox}
\end{table}

\noindent\textbf{Evaluation Protocol }
We report two settings. In MCSS, all modalities are available at training and inference, measuring full-modality fusion. In MISS, the model is trained with all modalities and evaluated on modality subsets at inference, measuring robustness when sensors are unavailable.

\noindent\textbf{Comparison Context }
Cross-backbone headline comparisons reflect both the method and the pretrained representation. We separately report internally trained controls with frozen MiT-B0/B2 and a common SegFormer decoder~\citep{xie2021segformer}, recipe, and checkpoint rule (Section~\ref{sec:controlled}). DINOv2-S was chosen on MUSES to keep the trainable budget comparable to MiT-B0 baselines, not to match backbone capacity. A matched DINOv2-B run reaches 80.16 versus 77.28 with DINOv2-S (Appendix~\ref{app:controls}).

\subsection{Modality-Complete Semantic Segmentation (MCSS) Results}
We first evaluate MCSS on MCubeS, MUSES, and DELIVER, where all modalities are available during training and inference.
As shown in Tabs.~\ref{tab:mss_mcubes}, \ref{tab:mss_muses}, and \ref{tab:mss_deliver},
FuseAdapt achieves the best reported mean performance among the compared methods on all three benchmarks with a small trainable budget.
On MCubeS, FuseAdapt obtains 57.62 mIoU with all four modalities (RADN) and 56.30 mean mIoU, compared with 53.29 for StitchFusion and 52.43 for MAGIC.
On DELIVER, it achieves 64.97 mean mIoU, outperforming StitchFusion (63.68) and Any2Seg (64.18).
On MUSES, it reaches 76.14 mean mIoU, improving over the prior reported results across all modality combinations.
These results indicate that adapting a frozen VFM with modality-specific paths and routed adaptation exchange provides effective full-modality fusion.

\begin{table}[!t]
\small
\renewcommand{\tabcolsep}{1pt}
\renewcommand{\arraystretch}{1.2}
\caption{Results in MCSS setting on DELIVER. The best results are highlighted in \textbf{bold}. (R: RGB; D: Depth; E: Event; L: LiDAR.)}
\centering
\begin{adjustbox}{width=0.8\columnwidth}
\begin{tabular}{l|c|c|cccccc|c}
\toprule
\multirow{2}{*}{Method} &\multirow{2}{*}{Backbone} &\multicolumn{1}{c|}{$\#$Trainable}& \multicolumn{6}{c|}{Modals} & \multirow{2}{*}{Mean}\\
\cmidrule{4-9}
 & &Params (M) & RD & RE & RL & RDE & RDL & RDEL & \\
\midrule
HRFuser~\citep{broedermann2023hrfuser}        & HRFuser-T & 31.61 & 51.88 & 42.22 & 43.13 & 51.83 & 52.72 & 52.97 & 49.13\\
CWSAM~\citep{pu2025classwise}                 & ViT-B     & 15.29 & 58.64 & -     & -     & 56.22 & -     & 55.43 & -\\
MLE-SAM~\citep{zhu2024customize}              & Hiera-B+  & 20.79 & 63.57 & -     & -     & 62.69 & -     & 64.08 & -\\
MemorySAM~\citep{liao2025memorysam}           & Hiera-B+  & 14.50 & 63.48 & -     & -     & 62.42 & -     & 65.38 & -\\
TokenFusion~\citep{wang2022multimodal}        & MiT-B2    & 26.01 & 60.25 & 45.63 & 53.01 & -     & -     & -     & -\\
CMX~\citep{zhang2023cmx}                      & MiT-B2    & 66.57 & 62.67 & 56.62 & 56.37 & -     & -     & -     & -\\
CMNeXt~\citep{zhang2023delivering}            & MiT-B2    & 58.73 & 63.58 & 57.48 & 58.04 & 64.44 & 65.50 & 66.30 & -\\
Zheng et al.~\citep{zheng2025reducing}        & MiT-B2    & 24.73 & 66.46 & 56.06 & -     & -     & -     & -     & -\\
MAGIC~\citep{zheng2024centering}              & MiT-B2    & 24.73 & 66.89 & 58.48 & 57.75 & 66.24 & \textbf{67.63} & 67.66 & -\\
StitchFusion~\citep{li2024stitchfusion}       & MiT-B2    & 26.50 & 65.75 & 57.44 & 58.03 & 66.03 & 66.65 & 68.18 & 63.68\\
Any2Seg~\citep{zheng2024learning}             & MiT-B2    & 24.73 & 67.12 & 57.83 & 57.92 & 66.75 & 67.20 & 68.25 & 64.18\\
OmniSegmentor~\citep{yin2025omnisegmentor}    & DFormer-L & 39.00 & 64.70 & 58.60 & 59.20 & 65.90 & 67.20 & 68.00 & 63.93\\
\midrule
\rowcolor{gray!20}
Ours                                         & ViT-B     & 14.01 & \textbf{67.33} & \textbf{59.62} & \textbf{60.11} & \textbf{67.16} & 67.29 & \textbf{68.33} & \textbf{64.97}\\
\bottomrule
\end{tabular}
\end{adjustbox}
\label{tab:mss_deliver}
\end{table}

\begin{table*}[!t]
\caption{Results in MISS setting on MUSES. All methods are trained with three modalities. The best results are highlighted in \textbf{bold}.}
\begin{adjustbox}{width=0.9\textwidth,center}
\centering
\begin{tabular}{l|c|c|ccccccc|c}
\toprule
\multirow{2}{*}{Method} &\multirow{2}{*}{Backbone} &\multicolumn{1}{c|}{$\#$Trainable}& \multicolumn{7}{c|}{Modals} & \multirow{2}{*}{Mean}\\
\cmidrule{4-10}
 & &Params (M) & F & E & L & FE & FL & EL & FEL \\
\midrule
CMX~\citep{zhang2023cmx}                        & MiT-B0   & 10.32 & 2.52  & 2.35  & 3.01  & 41.15 & 41.25 & 2.56  & 42.27 & 19.30 \\
CMNeXt~\citep{zhang2023delivering}              & MiT-B0   & 10.30 & 3.50  & 2.77  & 2.64  & 6.63  & 10.28 & 3.14  & 46.66 & 10.80 \\
MAGIC~\citep{zheng2024centering}                & MiT-B0   & 3.72  & 43.22 & 2.68  & 22.95 & 43.51 & 49.05 & 22.98 & 49.02 & 33.34 \\
Any2Seg~\citep{zheng2024learning}               & MiT-B0   & 3.72  & 44.40 & 3.17  & 22.33 & 44.51 & 49.96 & 22.63 & 50.00 & 33.86 \\
MLE-SAM~\citep{zhu2024customize}                & Hiera-B+ & 20.79 & 69.67 & 1.50  & 5.55  & 69.50 & 74.11 & 5.55  & 74.80 & 42.95 \\
Zheng et al.~\citep{zheng2025reducing}          & MiT-B0   & 3.72  & 60.33 & \textbf{33.15} & 42.59 & 47.19 & 53.47 & 39.59 & 47.89 & 46.32 \\
BiXFormer~\citep{chen2025bixformer}             & MiT-B0   & 17.70 & 65.76 & 16.88 & 39.63 & 58.42 & 59.77 & 49.89 & 65.81 & 50.88 \\
\midrule
\rowcolor{gray!20}
Ours                                          & ViT-S    & 3.29  & \textbf{71.09} & 30.16 & \textbf{46.59} & \textbf{73.24} & \textbf{76.72} & \textbf{54.04} & \textbf{77.28} & \textbf{61.30} \\
\bottomrule
\end{tabular}
\end{adjustbox}
\label{tab:miss_muses}
\end{table*}

\begin{figure*}[!t]
\centering
\includegraphics[width=0.9\linewidth]{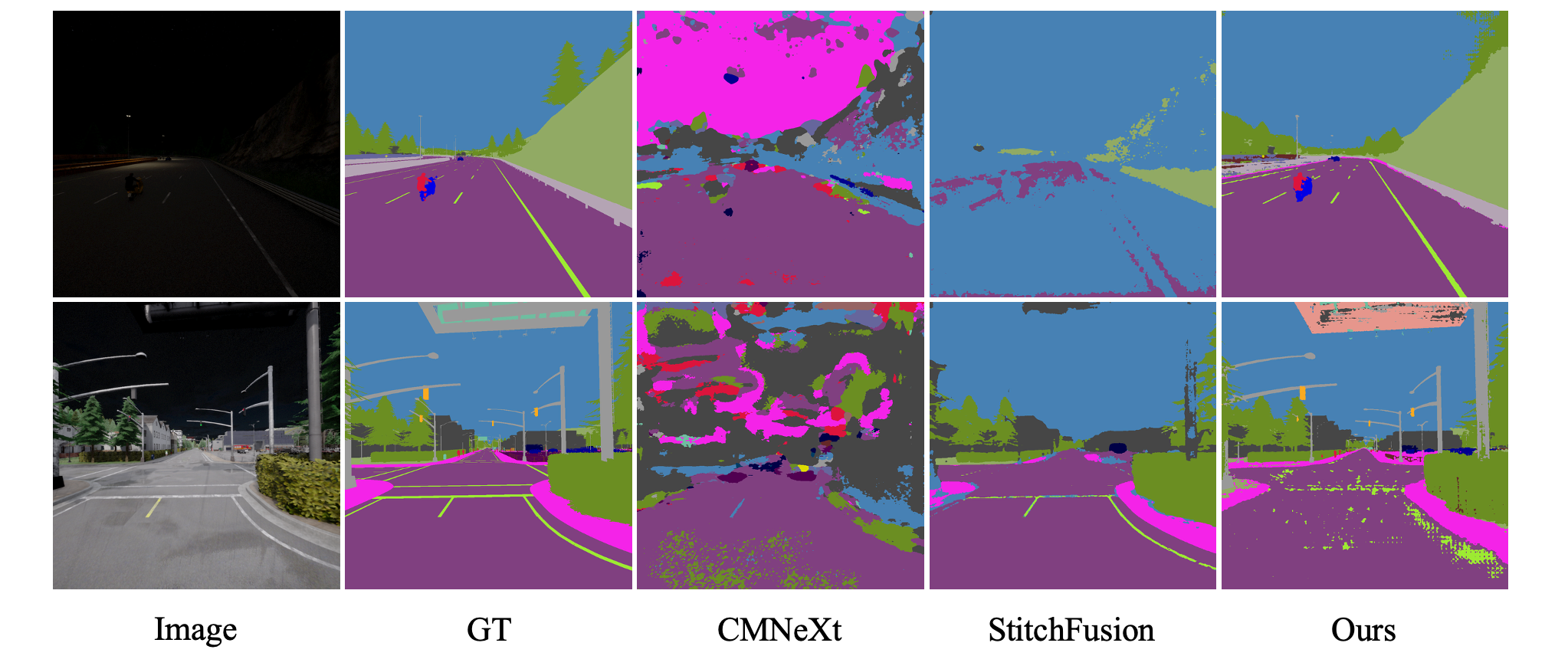}
\vspace{-2mm}
\caption{Qualitative results on DELIVER in MISS setting. The proposed method is compared with CMNeXt~\citep{zhang2023delivering} and StitchFusion~\citep{li2024stitchfusion}. The model is trained on four modalities $\{r,d,e,l\}$ and tested on $\{r\}$ (Row 1) and $\{r,l\}$ (Row 2).}
\label{fig:qual}
\end{figure*}

\subsection{Modality-Incomplete Semantic Segmentation (MISS) Results}
We next evaluate MISS, where models are trained with all modalities and tested on arbitrary modality subsets.
As shown in Tabs.~\ref{tab:miss_mcubes}, \ref{tab:miss_muses}, and \ref{tab:miss_deliver}, FuseAdapt improves performance across a broad range of missing-modality settings.
On MCubeS, FuseAdapt achieves 52.89 mean mIoU, outperforming MAGIC (38.00) and Zheng et al. (41.64).
On DELIVER, it obtains 56.22 mean mIoU, compared with 45.04 for Any2Seg, 44.85 for MAGIC++, and 41.98 for StitchFusion.
On MUSES, it reaches 61.30 mean mIoU with 3.29M trainable parameters, outperforming the reported baselines.
The gains across different modality subsets suggest that the learned representation remains useful under changing sensor availability.
Fig.~\ref{fig:qual} shows qualitative MISS results on DELIVER.
With only RGB (Row 1) or RGB+LiDAR (Row 2), FuseAdapt produces cleaner predictions and more coherent scene layouts than CMNeXt and StitchFusion, especially in challenging night or fog cases where baselines show larger region-level errors.

\subsection{Ablation Studies}
We ablate FuseAdapt and PMM under the same training recipe and also examine the effect of changing the frozen VFM backbone.

\noindent\textbf{FuseAdapt Variants (MCSS) }
Tabs.~\ref{tab:mss_mcubes} and~\ref{tab:mss_muses} compare variants within the same frozen-VFM framework.
Without FuseAdapt, the model uses naive aggregation of per-modality features and performs worse.
Shared FuseAdapt improves performance but uses one adaptation path for all modalities, limiting modality-specific correction.
Separate FuseAdapt uses modality-specific paths without routed exchange; it improves over Shared on MCubeS, while their MUSES DINOv2-S means are similar. Full FuseAdapt performs best in both tables.
This supports the contribution of both modality-specific adaptation and routed adaptation exchange.

\noindent\textbf{PMM Variants (MISS) }
Tab.~\ref{tab:miss_mcubes} evaluates PMM variants under MISS on MCubeS.
Removing the predictor or replacing it with a linear predictor reduces mean mIoU from 52.89 to 43.67 and 49.71, respectively.
This shows that the representation-level prediction objective is important, and that the query-based predictor is more effective than a simple linear prediction head.
Additional ablations, including FuseAdapt rank and scalability, are provided in the supplementary material.

\noindent\textbf{Additional VFMs }
To examine sensitivity to the frozen VFM choice, we keep the full pipeline fixed and replace DINOv2-B with EVA-CLIP-B~\citep{sun2023eva} and NeCo-B~\citep{pariza2024neco} on MCubeS under MCSS.
FuseAdapt achieves 56.02 mIoU with EVA-CLIP-B and 56.28 with NeCo-B, compared with 57.62 using DINOv2-B.
Although absolute performance depends on the chosen VFM, the method remains competitive across different frozen backbones, reducing dependence on a single pretrained model.

\begin{table*}[!t]
\renewcommand{\arraystretch}{1.2}
\caption{Results in MISS setting on MCubeS. All methods are trained with four modalities. The best results are highlighted in \textbf{bold}.}
\begin{adjustbox}{width=0.9\textwidth,center}
\centering
\setlength{\tabcolsep}{1pt}
\begin{tabular}{l|c|c|ccccccccccccccc|c}
\toprule
\multirow{2}{*}{Method} &\multirow{2}{*}{Backbone} &\multicolumn{1}{c|}{Trainable}& \multicolumn{15}{c|}{Modals} & \multirow{2}{*}{Mean}\\
\cmidrule{4-18}
 & &Params* & R & A & D & N & RA & RD & RN & AD & AN & DN & RAD & RAN & RDN & ADN & RADN \\
\midrule
CMNeXt~\citep{zhang2023delivering}          & MiT-B2 & 58.73 & 1.86 & 1.54 & 2.51 & 2.28 & 47.96 & 43.67 & 45.90 & 6.99 & 7.58 & 9.95 & 50.07 & 48.77 & 48.83 & 8.06 & 51.54 & 25.17 \\
MAGIC~\citep{zheng2024centering}            & MiT-B2 & 24.73 & 51.91 & 0.32 & 34.52 & 2.66 & 52.24 & 52.16 & 52.57 & 1.98 & 36.01 & 37.09 & 52.48 & 52.82 & 52.73 & 37.57 & 53.01 & 38.00 \\
Zheng et al.~\citep{zheng2025reducing}      & MiT-B2 & 24.73 & 46.21 & \textbf{37.66} & 40.49 & 40.56 & 43.93 & 43.11 & 43.73 & 40.10 & 39.19 & 40.72 & 41.78 & 41.53 & 42.21 & 40.63 & 42.73 & 41.64 \\
\midrule
Ours (wo pred.)                            & ViT-B  & 14.01 & 51.02 & 13.60 & 22.47 & 33.61 & 53.16 & 54.92 & 54.04 & 26.37 & 38.26 & 40.24 & 55.33 & 55.67 & 56.61 & 43.02 & 56.66 & 43.67 \\
Ours (linear pred.)                        & ViT-B  & 14.01 & 51.38 & 31.08 & 42.02 & 44.97 & 53.64 & 54.64 & 54.82 & 43.36 & 47.38 & 49.62 & 55.01 & 55.31 & 56.06 & 50.15 & 56.20 & 49.71 \\
\rowcolor{gray!20}
Ours                                       & ViT-B  & 14.01 & \textbf{54.96} & 37.60 & \textbf{47.14} & \textbf{49.12} & \textbf{56.17} & \textbf{57.23} & \textbf{56.29} & \textbf{47.96} & \textbf{51.65} & \textbf{52.74} & \textbf{57.19} & \textbf{56.80} & \textbf{57.58} & \textbf{53.30} & \textbf{57.62} & \textbf{52.89} \\
\bottomrule
\end{tabular}
\end{adjustbox}
\label{tab:miss_mcubes}
\end{table*}

\begin{table*}[!t]
\renewcommand{\arraystretch}{1.2}
\caption{Results in MISS setting on DELIVER. All methods are trained with four modalities. The best results are highlighted in \textbf{bold}.}
\begin{adjustbox}{width=0.9\textwidth,center}
\setlength{\tabcolsep}{1pt}
\centering
\begin{tabular}{l|c|c|ccccccccccccccc|c}
\toprule
\multirow{2}{*}{Method} &\multirow{2}{*}{Backbone} &\multicolumn{1}{c|}{$\#$Trainable}& \multicolumn{15}{c|}{Modals} & \multirow{2}{*}{Mean}\\
\cmidrule{4-18}
 & &Params (M) & R & D & E & L & RD & RE & RL & DE & DL & EL & RDE & RDL & REL & DEL & RDEL \\
\midrule
CMNeXt~\citep{zhang2023delivering}           & MiT-B2 & 58.73 & 22.50 & 50.59 & 3.16 & 2.86 & 66.33 & 22.92 & 22.50 & 50.80 & 50.83 & 3.15 & 66.27 & 66.38 & 22.92 & 50.98 & 66.33 & 37.90\\
GeminiFusion~\citep{jia2024geminifusion}     & MiT-B2 & 42.18 & 15.89 & 54.73 & 1.70 & 1.70 & 66.93 & 16.24 & 15.80 & 54.83 & 54.76 & 1.70 & 66.92 & 66.93 & 16.18 & 54.86 & 66.92 & 37.07\\
MAGIC~\citep{zheng2024centering}             & MiT-B2 & 24.73 & 42.72 & 58.39 & 1.90 & 1.62 & 66.10 & 42.79 & 42.72 & 58.44 & 58.39 & 1.90 & 66.11 & 66.10 & 42.80 & 58.44 & 66.10 & 44.97\\
MAGIC++~\citep{zheng2024magic++}             & MiT-B2 & 27.51 & 41.10 & 58.12 & 2.14 & 1.64 & 67.33 & 41.13 & 41.13 & 58.32 & 58.12 & 2.15 & 67.35 & 67.33 & 41.17 & 58.31 & 67.34 & 44.85\\
Any2Seg~\citep{zheng2024learning}            & MiT-B2 & 24.73 & 39.02 & \textbf{60.11} & 2.07 & 0.31 & 68.21 & 39.11 & 39.04 & \textbf{60.92} & \textbf{60.15} & 1.99 & 68.24 & 68.22 & 39.06 & \textbf{60.95} & 68.25 & 45.04\\
StitchFusion~\citep{li2024stitchfusion}      & MiT-B2 & 26.50 & 30.93 & 55.44 & 1.87 & 1.59 & 68.22 & 31.03 & 33.55 & 55.76 & 55.41 & 1.87 & 68.23 & 68.21 & 33.66 & 55.71 & 68.20 & 41.98\\
Zheng et al.~\citep{zheng2025reducing}       & MiT-B2 & 24.73 & 55.56 & 57.74 & \textbf{32.40} & \textbf{34.26} & 59.69 & 47.70 & 47.78 & 48.99 & 50.26 & 34.37 & 53.42 & 54.12 & 45.24 & 47.33 & 51.65 & 48.03\\
\midrule
\rowcolor{gray!20}
Ours                                        & ViT-B  & 14.01 & \textbf{58.52} & 59.48 & 30.73 & 27.67 & \textbf{68.93} & \textbf{59.43} & \textbf{58.08} & 60.39 & 58.89 & \textbf{37.72} & \textbf{68.90} & \textbf{68.42} & \textbf{58.58} & 59.22 & \textbf{68.33} & \textbf{56.22} \\
\bottomrule
\end{tabular}
\end{adjustbox}
\label{tab:miss_deliver}
\end{table*}

\subsection{Controlled Comparisons}
\label{sec:controlled}
\textbf{Backbone and recipe controls.} On MUSES with frozen MiT-B0, FuseAdapt obtains 57.54 mIoU versus 49.47 for Deep Prompt, 53.56 for Shared LoRA, and 54.80 for Adapter. These internal controls use the same decoder, split, augmentation, optimizer, schedule, PMM, and checkpoint rule, with 2.95/2.81/2.95/2.94M trainable parameters, respectively. External MiT-B0 references MAGIC and Any2Seg obtain 49.02/50.00. With MiT-B2, FuseAdapt obtains 70.60, compared with 58.28/55.36/53.40/59.36 for CMNeXt/MAGIC/StitchFusion/CHARM. External results share the backbone family but are not all recipe-matched; Appendix~\ref{app:controls} gives definitions and protocols.

\textbf{Alternative fusion mechanisms.} Table~\ref{tab:fusion_full} further compares fusion designs on MCubeS using DINOv2-B and the same segmentation decoder. Frozen feature averaging and concatenation obtain 51.16 and 52.42 mIoU, while token-attention fusion obtains 52.79. Late-logit fusion with private low-rank adaptation reaches 54.55, compared with 57.62 for FuseAdapt. These controls distinguish adaptation-space exchange from simply combining frozen representations or independently adapted predictions. Separate adaptation without routing reaches 56.09, narrowing the gap but remaining below the full model.

\textbf{Where should information cross modalities?} Dense-feature routing keeps Q/V private but mixes complete FFN features; All-path routed LoRA routes Q/V and FFN residuals. Both retain the backbone and decoder of the corresponding setting. FuseAdapt outperforms these alternatives on MCubeS and MUSES (Table~\ref{tab:infoflow}), supporting the separation of private attention adaptation and residual-only exchange. These are internal mechanism controls, not reproductions of complete prior systems.
\begin{table}[t]
\centering\small
\caption{Full-modality mIoU for information-flow variants (seed 3407).}
\label{tab:infoflow}
\begin{tabular}{lrr}
\toprule
Variant & MCubeS / DINOv2-B & MUSES / MiT-B0\\
\midrule
Shared & 54.69 & 53.56\\
Separate & 56.09 & 54.31\\
Dense-feature routing & 55.08 & 55.94\\
All-path routed LoRA & 55.74 & 56.17\\
FuseAdapt & \textbf{57.62} & \textbf{57.54}\\
\bottomrule
\end{tabular}
\end{table}

\textbf{Prediction versus generic distillation.} With the same backbone, masking, EMA teacher, segmentation supervision, and schedule, output-logit KD and fused-feature KD obtain 49.84 and 50.57 MISS mean mIoU on MCubeS, versus 52.89 for PMM. PMM improves output KD in all 15 nonempty subsets and fused-feature KD in 14/15. Removing EMA from the query predictor gives 50.59; the original per-token regressor gives 49.71. These controls support the target-conditioned prediction objective beyond generic full-to-partial matching. Table~\ref{tab:kd_summary} summarizes the objectives; Appendix~\ref{app:kd} gives every subset.

\begin{table}[!t]
\centering\small
\caption{MCubeS MISS mean mIoU over all 15 nonempty subsets. Only the auxiliary objective/target differs; full results are in Table~\ref{tab:kd_full}.}
\label{tab:kd_summary}
\begin{tabular}{lr@{\qquad}lr}
\toprule Objective & Mean & Objective & Mean\\\midrule
Dropout only & 43.67 & Query without EMA & 50.59\\
Per-token regression & 49.71 & Global-feature KD & 50.57\\
Output KD & 49.84 & Full PMM & \textbf{52.89}\\\bottomrule
\end{tabular}
\end{table}

\subsection{Robustness and Deployment Cost}
\textbf{Scope of robustness.} On DELIVER's ten predefined weather/degradation conditions, FuseAdapt obtains 67.36 mean mIoU versus 67.19 for Any2Seg, leading in eight conditions but trailing at night and under-exposure. In a separate MCubeS latent-prediction test, shifting the visible sensors together by 16 pixels relative to the masked target reduces cosine similarity by 0.00051 for the query predictor versus 0.01636 for per-token regression. The linear model has higher clean cosine; the evidence concerns lower shift sensitivity, not universally better reconstruction or arbitrary sensor misalignment. Appendix~\ref{app:robustness} provides both protocols and full tables.

\textbf{Full-context target reliability.} A full-input EMA target is useful on average but is not a per-sample oracle. Across 1,428 MCubeS image/subset pairs, comprising 102 images and 14 proper subsets, full-input predictions improve per-image mIoU by 4.44 on average and outperform partial-input predictions in 81.72\% of pairs. PMM nevertheless uses detached latent representations, rather than segmentation pseudo-labels, as supervision. This analysis characterizes the benefit and limitation of the full-context assumption: additional sensors usually provide useful context under the evaluated distribution, but can also introduce misleading information. See Appendix~\ref{app:teacher} for details.

\textbf{End-to-end efficiency.} On an idle H100, at $1024^2$, batch 1, FP32, with 20 warm-ups and 100 timed runs, three-modality MiT-B0 FuseAdapt takes 21.05 ms versus 19.63 for CMNeXt and 26.69 for StitchFusion. It retains 3.91M unique deployed parameters and removes both PMM components. Parameter efficiency does not imply lowest latency or memory: Any2Seg and MAGIC are faster in this profile. Appendix~\ref{app:efficiency} reports all measured costs and identifies reconstructed inference graphs.

\section{Discussion and Conclusion}
\label{sec:discussion_conclusion}
FuseAdapt separates modality-private attention adaptation from cross-modal exchange through FFN low-rank residuals. PMM complements this design with target-conditioned missing-modality latent prediction. Same-backbone controls support both choices, with improvements across the evaluated MiT scales.

\textbf{Limitations and deployment scope.} Absolute accuracy depends on the frozen representation and its biases; cross-backbone comparisons alone cannot isolate our contribution. Inference cost still grows with the number of sensor streams. Routing and averaging require a common projected grid, and the shift experiment measures latent prediction sensitivity, not robustness to arbitrary calibration, field-of-view, or temporal errors. DELIVER covers predefined adverse conditions, but neither those results nor a full-context EMA target guarantees safe behavior under correlated or out-of-distribution sensor failures, including device-level ones~\citep{yuan2025large,yuan2025enhanced}. Deployment in robotics or driving requires application-specific testing, failure detection~\citep{chen2025autodiscovery}, uncertainty estimation~\citep{chen2026hypospace}, and validated fallback mechanisms.

\clearpage
\begin{ack}
This research is supported by the Ministry of Education, Singapore, under its
MOE AcRF Tier 3 Grant (MOE-MOET32022-0001). The authors declare no competing interests.
\end{ack}

\bibliography{references}
\bibliographystyle{plainnat}

\appendix

\newpage

\section{Controlled Comparisons and Experimental Protocols}
\label{app:controls}
We separate external same-backbone references from controls trained within our framework. For MUSES MiT-B0/B2, internal controls use a frozen encoder, SegFormer decoder, the same split, augmentations, optimizer, 200-epoch schedule, $1024\times1024$ inputs, rank-4 adaptation where applicable, EMA-query PMM, effective global batch size 16, and the best full-modality validation checkpoint selected independently for each run. Seeds are 3407, 3408, and 3409 for repetitions and 3407 for individual controls. These experiments test backbone generality on the evaluated hierarchical MiT scales, not arbitrary architectures.

Deep Prompt inserts 64 modality-specific learnable prompts into every Transformer block. Shared LoRA applies one rank-4 Q/V and FFN update shared across sensors. Adapter inserts modality-specific width-32 residual bottlenecks in stages 2--4. Separate LoRA uses private Q/V and FFN residual branches without cross-modal routing. Full FuseAdapt adds receiver-conditioned FFN residual exchange. All variants average the resulting visible features before the shared decoder.

\begin{table}[htbp]
\centering\small
\caption{MUSES full-modality mIoU with MiT-B0. External references share the backbone family; internal controls additionally share the decoder and recipe. Trainable parameter counts include optimized components of the internal setup.}
\label{tab:peft_full}
\begin{tabular}{llrr}
\toprule Method & Comparison & Trainable (M) & mIoU\\\midrule
MAGIC & External & 3.72 & 49.02\\
Any2Seg & External & 3.72 & 50.00\\\midrule
Deep Prompt & Internal & 2.95 & 49.47\\
Shared LoRA & Internal & 2.81 & 53.56\\
Adapter & Internal & 2.95 & 54.80\\
FuseAdapt & Internal & 2.94 & \textbf{57.54}\\\bottomrule
\end{tabular}
\end{table}

\begin{table}[htbp]
\centering\small
\caption{MUSES full-modality mIoU using MiT-B2. Any2Seg has no reported MUSES/MiT-B2 result and is omitted.}
\label{tab:mit_b2_external}
\begin{tabular}{lrlr}
\toprule External method & mIoU & Internal variant (seed 3407) & mIoU\\\midrule
CMNeXt & 58.28 & Shared & 65.38\\
MAGIC & 55.36 & Separate & 66.96\\
StitchFusion & 53.40 & FuseAdapt & \textbf{70.60}\\
CHARM~\citep{wen2026charm} & 59.36 & & \\\bottomrule
\end{tabular}
\end{table}

\paragraph{DINOv2 capacity.}
Under the same MUSES split, seed 3407, 200 epochs, rank 4, EMA-query PMM, optimizer schedule, and effective batch size 16, DINOv2-S/B achieve 77.28/80.16 full-modality best-validation mIoU. The 2.88-point difference confirms the contribution of backbone capacity. DINOv2-S was selected to keep the original 3.29M trainable budget comparable to MAGIC/Any2Seg (3.72M each); it does not match their backbone family.

\paragraph{Fusion controls on MCubeS.}
All rows in Table~\ref{tab:fusion_full} use DINOv2-B and a shared segmentation decoder. Frozen mean averages frozen per-modality features without LoRA. Frozen concatenation concatenates those features and projects them through a trainable $1\times1$ layer at each scale. Token-attention fusion uses eight-head attention across sensor tokens at each common-grid location. Late-logit fusion uses private rank-8 Q/V and FFN updates without routing, decodes each stream independently with the shared head, and averages logits. Shared and Separate use shared or private low-rank branches, respectively, followed by feature averaging.

\begin{table}[htbp]
\centering\small
\caption{MCubeS, all four modalities (RADN): fusion/adaptation mIoU.}
\label{tab:fusion_full}
\begin{tabular}{lr}
\toprule Method & RADN mIoU\\\midrule
Frozen mean & 51.16\\
Frozen concatenation & 52.42\\
Token-attention fusion & 52.79\\
Late-logit fusion & 54.55\\
Shared & 54.69\\
Separate & 56.09\\
FuseAdapt & \textbf{57.62}\\\bottomrule
\end{tabular}
\end{table}

\paragraph{Information-flow alternatives.}
All-path routed LoRA routes the low-rank Q/V residuals as well as the FFN residuals, testing whether attention adaptation should also mix across sensors. Dense-feature routing preserves private Q/V but exchanges complete FFN features, testing whether the low-rank residual restriction matters. The MUSES MiT-B0 controls share rank, PMM, training recipe, effective batch size, seed, trainable parameter count, and checkpoint rule. Table~\ref{tab:infoflow} also reports the corresponding MCubeS/DINOv2-B results. These controlled alternatives isolate architectural choices rather than reproducing StitchFusion or MLE-SAM in full.

\section{Multi-Seed Reliability}
\label{app:seeds}
Table~\ref{tab:seeds_full} reports all runs in the MUSES MiT comparison. Each run is trained for the same budget and evaluated using its validation-selected full-modality checkpoint. For values $x_1,x_2,x_3$, we report $\bar{x}=\frac{1}{3}\sum_i x_i$ and sample standard deviation $s=\sqrt{\sum_i(x_i-\bar{x})^2/(3-1)}$. This measures run-to-run variation under a fixed split and recipe; it is not a standard error or confidence interval. We do not infer formal significance from three seeds.
\begin{table}[htbp]
\centering\small
\caption{MUSES full-modality mIoU for each random seed and mean $\pm$ sample SD.}
\label{tab:seeds_full}
\begin{tabular}{llrrrr}
\toprule Backbone & Variant & 3407 & 3408 & 3409 & Mean $\pm$ SD\\\midrule
MiT-B0 & Shared & 53.56 & 54.74 & 54.09 & $54.13\pm0.59$\\
& Separate & 54.31 & 56.34 & 56.94 & $55.86\pm1.38$\\
& Full & 57.54 & 57.77 & 57.48 & $57.60\pm0.15$\\\midrule
MiT-B2 & Shared & 65.38 & 65.69 & 66.32 & $65.80\pm0.48$\\
& Separate & 66.96 & 68.29 & 66.95 & $67.40\pm0.77$\\
& Full & 70.60 & 70.67 & 70.58 & $70.62\pm0.05$\\\bottomrule
\end{tabular}
\end{table}
Full exceeds Shared in all six paired comparisons: by 3.98/3.03/3.39 on MiT-B0 and 5.22/4.98/4.26 on MiT-B2. The mean improvements over Shared are 3.47/4.82, and over Separate 1.74/3.22. In a separate three-seed fixed-checkpoint evaluation, the parameter-matched Shared control and Full obtain $54.52\pm0.29$ and $57.41\pm0.27$ on MCubeS, and $76.63\pm0.43$ and $77.17\pm0.38$ on MUSES with DINOv2-S. The latter gain of 0.54 is modest relative to variation; the clearer MiT margins should not be generalized to all backbones. A stronger Shared representation may leave less room for routing, but this is an interpretation rather than a causal result.

\section{PMM Objective Controls and Teacher Reliability}
\label{app:kd}
The MCubeS MISS comparisons keep the backbone, decoder, modality masking, segmentation supervision, and training schedule fixed. Output KD matches partial-input student logits to full-context EMA teacher logits through pixelwise KL divergence. Global-feature KD matches the student's fused representation to the full-context EMA fused representation using cosine loss. Neither identifies a missing target sensor. Per-token regression instead predicts its latent tokens at the same spatial indices. Query without EMA uses the same query predictor as PMM but a detached online full-context target. Full PMM uses an EMA target and target-specific query cross-attention. An EMA teacher without a matching loss would not provide student supervision.

\begin{table}[htbp]
\centering\small
\caption{MCubeS MISS mIoU for all objective controls. Mean averages all 15 nonempty subsets, including RADN. R/A/D/N denote RGB/AoLP/DoLP/NIR.}
\label{tab:kd_full}
\begin{tabular}{lrrrrrr}
\toprule
Subset & Dropout & Output KD & Linear & Query, no EMA & Feature KD & PMM\\\midrule
R & 51.02 & 51.40 & 51.38 & 53.97 & 53.46 & \textbf{54.96}\\
A & 13.60 & 34.92 & 31.08 & 37.99 & \textbf{38.66} & 37.60\\
D & 22.47 & 46.13 & 42.02 & 44.75 & 43.70 & \textbf{47.14}\\
N & 33.61 & 45.79 & 44.97 & 46.67 & 45.65 & \textbf{49.12}\\
RA & 53.16 & 52.08 & 53.64 & 55.10 & 54.27 & \textbf{56.17}\\
RD & 54.92 & 53.99 & 54.64 & 54.57 & 55.06 & \textbf{57.23}\\
RN & 54.04 & 52.62 & 54.82 & 53.06 & 53.64 & \textbf{56.29}\\
AD & 26.37 & 45.39 & 43.36 & 46.35 & 44.64 & \textbf{47.96}\\
AN & 38.26 & 48.33 & 47.38 & 48.88 & 48.66 & \textbf{51.65}\\
DN & 40.24 & 49.63 & 49.62 & 49.67 & 49.61 & \textbf{52.74}\\
RAD & 55.33 & 53.70 & 55.01 & 54.86 & 55.40 & \textbf{57.19}\\
RAN & 55.67 & 53.36 & 55.31 & 53.63 & 54.09 & \textbf{56.80}\\
RDN & 56.61 & 54.71 & 56.06 & 54.52 & 55.47 & \textbf{57.58}\\
ADN & 43.02 & 50.56 & 50.15 & 50.32 & 50.75 & \textbf{53.30}\\
RADN & 56.66 & 55.00 & 56.20 & 54.58 & 55.55 & \textbf{57.62}\\\midrule
Mean & 43.67 & 49.84 & 49.71 & 50.59 & 50.57 & \textbf{52.89}\\\bottomrule
\end{tabular}
\end{table}
PMM exceeds output KD by 3.05 mean mIoU and fused-feature KD by 2.32. It improves the former in all 15 subsets and the latter in 14/15. Linear prediction improves dropout-only training by 6.04; query prediction improves the linear control by 3.18; EMA improves the no-EMA query by 2.30. These results support modality-conditioned latent prediction beyond a generic consistency objective, without claiming masking or EMA as new mechanisms.

\subsection{Full-context targets are an average prior}
\label{app:teacher}
Teacher and student share the same frozen initialization and architecture. Only student adaptation parameters receive gradients, and teacher adaptation parameters follow the EMA update with coefficient 0.999. PMM uses detached teacher representations rather than segmentation pseudo-labels. Full-modality input does not guarantee better predictions when a sensor is degraded. Over 1,428 MCubeS image/subset pairs (102 images and 14 proper subsets), full-input predictions improve per-image mIoU by 4.44 on average and are better in 81.72\% of pairs. Thus full context is useful on average under this distribution, but it is not a per-sample oracle. Reliability-aware target weighting under sensor corruption remains future work.

\section{Degradation and Spatial-Shift Evaluation}
\label{app:robustness}
\paragraph{Degraded but present modalities.}
Table~\ref{tab:degradation} evaluates all four DELIVER modalities (R-D-E-L) on the benchmark's predefined weather and sensor-degradation subsets. MB/OE/UE/LJ/EL denote motion blur, over-exposure, under-exposure, LiDAR jitter, and event low resolution. This is not an additional arbitrary-corruption benchmark. FuseAdapt leads eight conditions and the mean, while Any2Seg leads Night and UE.
\begin{table}[htbp]
\centering\small
\caption{DELIVER, four present modalities: mIoU by predefined condition.}
\label{tab:degradation}
\begin{tabular}{lrrrr}
\toprule Condition & HRFuser & CMNeXt & Any2Seg & FuseAdapt\\\midrule
Cloudy & 56.20 & 68.70 & 69.19 & \textbf{69.22}\\
Foggy & 52.39 & 65.67 & 67.64 & \textbf{67.93}\\
Night & 49.85 & 62.46 & \textbf{67.25} & 66.44\\
Rainy & 52.53 & 67.50 & 67.00 & \textbf{68.76}\\
Sunny & 54.02 & 66.57 & 68.25 & \textbf{68.93}\\
MB & 49.44 & 62.91 & 66.23 & \textbf{66.39}\\
OE & 46.31 & 64.59 & 64.89 & \textbf{67.21}\\
UE & 46.92 & 60.00 & \textbf{65.69} & 61.39\\
LJ & 53.94 & 65.92 & 69.05 & \textbf{69.08}\\
EL & 52.72 & 65.48 & 66.73 & \textbf{68.23}\\\midrule
Mean & 51.43 & 64.98 & 67.19 & \textbf{67.36}\\\bottomrule
\end{tabular}
\end{table}

\paragraph{Latent prediction under spatial shifts.}
We mask each MCubeS target sensor in turn, then jointly shift the other three visible sensors by 0/2/4/8/16 input pixels relative to the clean full-context EMA target. Visible--visible alignment is preserved while target--visible correspondence changes. We compute channel-wise cosine between predicted target tokens and their teacher counterparts, averaged over tokens, 102 images, four target modalities, and three seeds. At 0 pixels the target is still missing and visible inputs are unshifted.
\begin{table}[htbp]
\centering\small
\caption{MCubeS predicted/teacher latent cosine under target--visible shifts. Final column reports mean drop $\pm$ SD as reported across the three seeds.}
\label{tab:shift_full}
\begin{tabular}{lrrrrrl}
\toprule Predictor & 0 px & 2 px & 4 px & 8 px & 16 px & Drop, 0--16 px\\\midrule
Linear & .86968 & .86766 & .86583 & .86136 & .85332 & $.01636\pm.00323$\\
Query & .74025 & .74012 & .74008 & .74006 & .73974 & $.00051\pm.00029$\\\bottomrule
\end{tabular}
\end{table}
Query prediction has the smaller drop in all 12 seed--target comparisons at every nonzero shift; its 16-pixel drop is approximately 32 times smaller. Linear prediction has higher absolute clean cosine. The experiment measures sensitivity of latent prediction to these translations, not general reconstruction superiority or end-to-end robustness to arbitrary native token lengths, fields of view, or temporal offsets.

\section{End-to-End Inference Efficiency}
\label{app:efficiency}
We profile all methods on the same idle H100 at $1024\times1024$, batch size 1, FP32, with 20 warm-up iterations and 100 measured runs. FuseAdapt, CMNeXt, and StitchFusion use MiT-B0 and 256-dimensional segmentation decoders. Any2Seg and MAGIC use MiT-B0 with their paper-specific heads. Deployed parameters count unique parameters retained for inference, counting shared backbones once and removing PMM's EMA teacher and query predictor. An asterisk identifies an inference graph reconstructed from its paper because executable code was unavailable at the time of profiling; the dagger denotes the released DFormer-L+LightHam proxy for OmniSegmentor.

\begin{table}[htbp]
\centering\small
\caption{End-to-end inference cost under the common H100 protocol. Rows list one, two, or three input modalities. Latency is ms, throughput is images/s, memory is peak GiB, and parameters are deployed millions.}
\label{tab:efficiency_full}
\begin{tabular}{llrrrr}
\toprule Method & Modalities & Latency & Throughput & Memory & Params\\\midrule
CMNeXt & 1 & 7.30 & 136.96 & 0.70 & 3.72\\
& 2 & 18.45 & 54.19 & 0.74 & 10.29\\
& 3 & 19.63 & 50.93 & 0.75 & 10.30\\\midrule
StitchFusion & 1 & 8.32 & 120.26 & 0.70 & 4.26\\
& 2 & 16.41 & 60.93 & 0.72 & 4.39\\
& 3 & 26.69 & 37.47 & 0.73 & 4.56\\\midrule
Any2Seg* & 1 & 6.53 & 153.08 & 0.60 & 3.72\\
& 2 & 10.26 & 97.43 & 1.16 & 3.72\\
& 3 & 14.69 & 68.09 & 1.71 & 3.72\\\midrule
MAGIC* & 1 & 6.49 & 154.16 & 0.60 & 3.72\\
& 2 & 10.25 & 97.53 & 1.16 & 3.79\\
& 3 & 14.72 & 67.92 & 1.71 & 3.85\\\midrule
FuseAdapt & 1 & 9.40 & 106.33 & 0.73 & 3.78\\
& 2 & 15.87 & 63.02 & 1.18 & 3.84\\
& 3 & 21.05 & 47.50 & 1.75 & 3.91\\\midrule
MLE-SAM* & 1 & 33.91 & 29.49 & 0.95 & 89.04\\
& 2 & 61.95 & 16.14 & 1.06 & 89.04\\
& 3 & 89.72 & 11.15 & 1.17 & 89.04\\\midrule
MemorySAM & 1 & 35.56 & 28.12 & 0.87 & 84.08\\
& 2 & 78.39 & 12.76 & 1.21 & 84.08\\
& 3 & 125.23 & 7.99 & 1.43 & 84.08\\\midrule
OmniSegmentor$^\dagger$ & 2 & 45.48 & 21.99 & 0.93 & 39.02\\\bottomrule
\end{tabular}
\end{table}
FuseAdapt's three-input latency is close to CMNeXt and below StitchFusion in this profile; its unique deployed parameter count is smaller than both. However, Any2Seg/MAGIC are faster, and CMNeXt/StitchFusion use less peak memory. Reconstructed graphs and the OmniSegmentor proxy should not be interpreted as official implementation benchmarks. We omit traced FLOPs because unsupported routing, normalization, and interpolation operations cause method-dependent undercounting. Small adaptation-module overhead should not be confused with the total cost of each additional frozen-encoder stream.
\clearpage

\section{Methodological Justification}

\subsection{Rationale for Using Modality-Specific LoRA in Self-Attention Adaptation.}

To examine whether self-attention should apply a fused LoRA module across modalities or maintain modality-specific adaptation paths, we analyze the learned LoRA updates $\Delta W$  across both the q and v projections for each modality on two datasets, DELIVER and MCubeS.
For each projection, we also evaluate shallow, mid, and deep transformer layers and report cosine-similarity matrices.

\begin{figure}[t]
	\centering
	\includegraphics[width=1.0\linewidth]{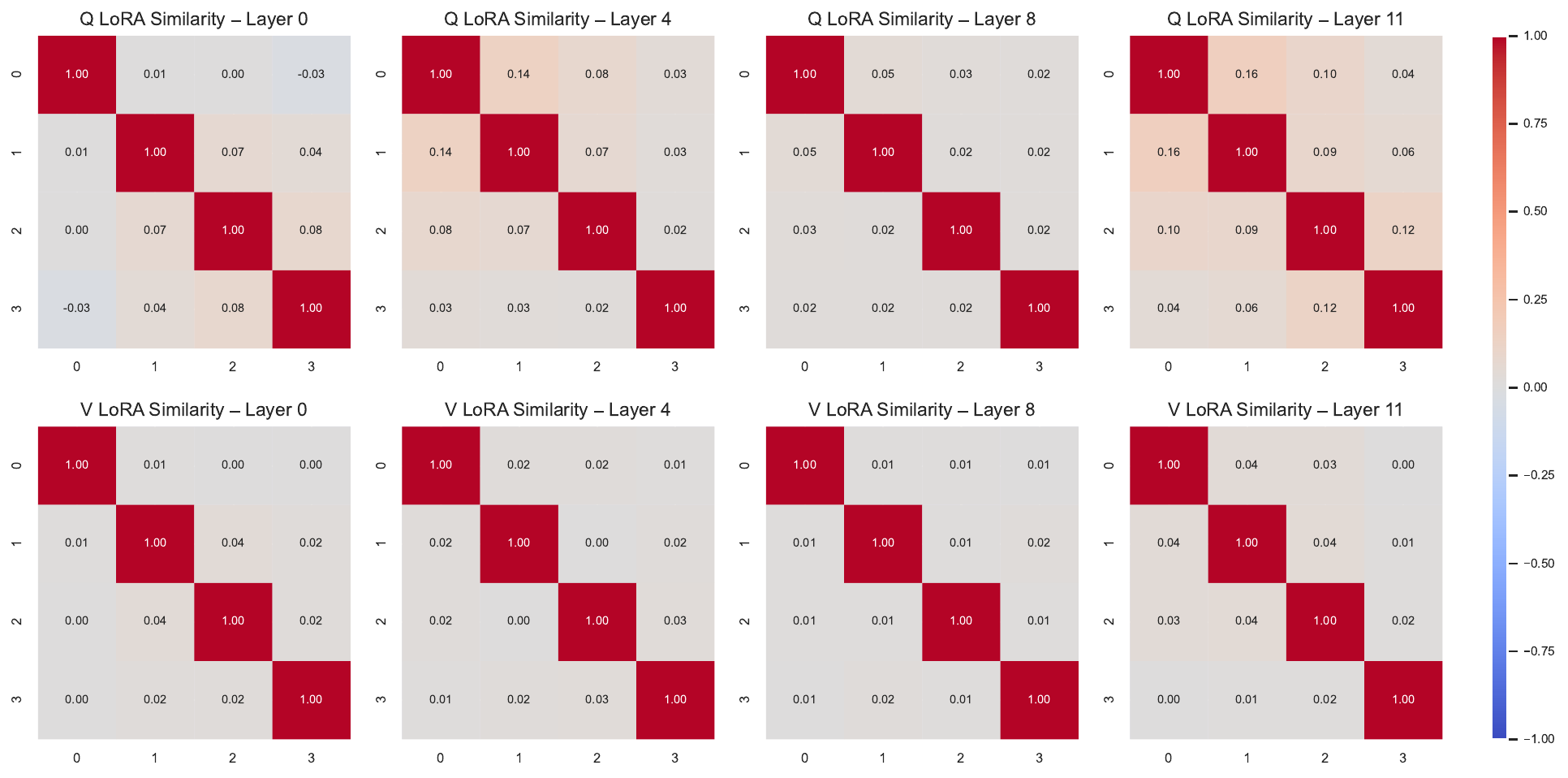}
	\caption{Cosine similarity matrices of modality-specific LoRA updates for the Q-projection (top row) and V-projection (bottom row) across different transformer layers. Each heatmap shows the pairwise cosine similarity between the LoRA updates $\Delta W$ of the four modalities of DELIVER. Across all layers, different modalities exhibit consistently low pairwise cosine similarity, indicating that the LoRA updates for each modality occupy distinct and nearly orthogonal subspaces. The indices 0, 1, 2, and 3 correspond to the four input modalities.}
	\label{fig:lora_deliver}
\end{figure}

\begin{figure}[t]
	\centering
	\includegraphics[width=1.0\linewidth]{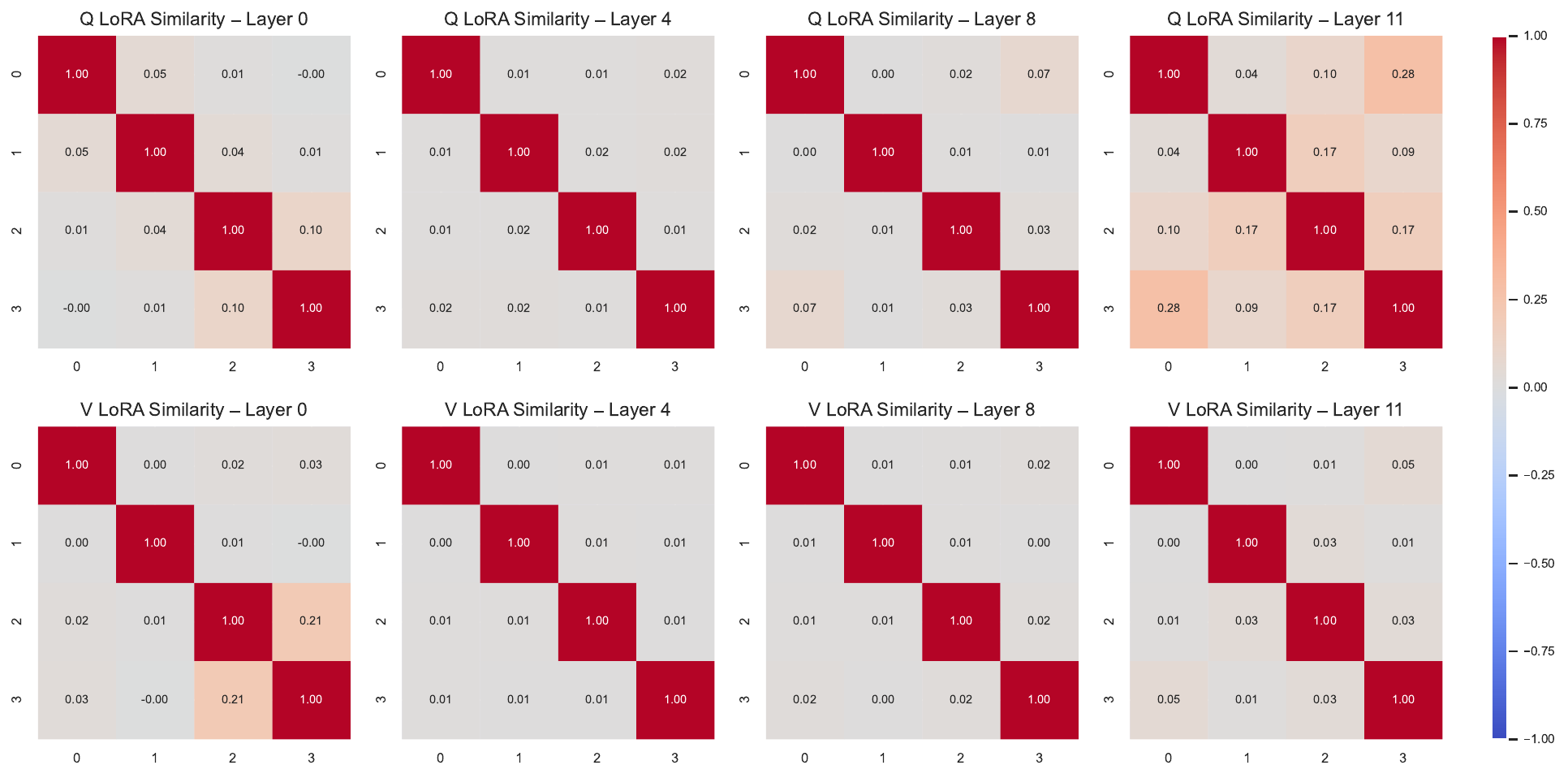}
	\caption{Cosine similarity matrices of modality-specific LoRA updates for the Q-projection (top row) and V-projection (bottom row) across different transformer layers. Each heatmap shows the pairwise cosine similarity between the LoRA updates $\Delta W$ of the four modalities in MCubeS. As with DELIVER, the cross-modal LoRA directions remain consistently low in similarity across all layers. The indices 0, 1, 2, and 3 correspond to the four input modalities.}
	\label{fig:lora_mcubes}
\end{figure}

As shown in Fig.~\ref{fig:lora_deliver} and Fig.~\ref{fig:lora_mcubes}, across all layers and for both projections, the LoRA updates of different modalities exhibit consistently low pairwise cosine similarity, forming near-orthogonal subspaces rather than converging toward a shared direction. This indicates that each modality learns distinct and highly specialized correction directions within the attention mechanism.

Low pairwise similarity supports distinct sensor-specific correction directions, but does not by itself establish that routing them is harmful. The all-path routing comparison in Table~\ref{tab:infoflow} tests that choice directly. Keeping Q/V private retains each sensor's attention rule while the FFN residual path permits cross-modal communication.

\subsection{Comparison with Mixture-of-Experts (MoE)}
Each FuseAdapt low-rank branch has a fixed source-sensor identity, while receiver-conditioned routing weights vary by token and input. All available sensor branches participate without top-$k$ selection or a load-balancing loss; absent sensors are masked. This distinguishes branch assignment from dynamic communication and does not imply fixed mixing weights. A sparse MoE-style LoRA control obtains 54.94 versus 57.62 mIoU on MCubeS. The all-path and dense-feature controls in Table~\ref{tab:infoflow} more directly test which activations should cross sensor streams.

\subsection{Details of PMM.}
\label{app:pmm_details}
After the full-input warm-up, with probability $0.5$ we retain all modalities; otherwise we draw independent Bernoulli-$0.5$ visibility masks and enforce a nonempty visible set per sample. A target is sampled only when at least one modality is missing. The teacher always receives the full input; its designated target-modality representation is detached. Modality-specific query sets attend to visible features and produce the same number of tokens as the target. For full-input samples, the missing-target PMM term is zero. Inference uses the actually available subset: all modalities for MCSS and the evaluated subset for MISS. The teacher and predictor are discarded.

\subsection{Class-balanced OHEM Cross-Entropy Loss.}
To better handle class imbalance and focus optimization on informative pixels, we adopt a weighted Online Hard Example Mining (OHEM) strategy with per-batch adaptive class weights, based on the vanilla OHEM implemented in DELIVER. Given the prediction $\mathbf{P} \in \mathbb{R}^{B \times C \times H \times W}$ and ground-truth labels $\mathbf{Y} \in \mathbb{R}^{B \times H \times W}$, the pixel-wise cross-entropy loss is first computed as:

\begin{equation}
	\mathcal{L}_{\text{CE}}(i) = - \sum_{c=1}^{C} w_c \cdot \mathbf{1}[y_i = c] \cdot \log P_{i,c},
\end{equation}

where $w_c$ denotes the class-specific weight for class $c$, computed per batch using the inverse frequency of observed labels, clipped to a maximum value of $w_{\text{max}}$:

\begin{equation}
	w_c =
	\begin{cases}
		\min \left( \dfrac{1}{f_c + \epsilon} \bigg/ \min\limits_{c'} \dfrac{1}{f_{c'} + \epsilon},\; w_{\text{max}} \right), & \text{if } f_c > 0 \\
		w_{\text{default}}, & \text{otherwise}
	\end{cases}
\end{equation}

where $f_c$ is the pixel count of class $c$ in the current batch and $w_{\text{default}}$ is the fallback weight for absent classes. $w_{\text{max}}$ is set to 10, and $w_{\text{default}}$ is set to 1.

To prioritize informative samples, we follow OHEM and select pixels with loss greater than a threshold $\tau$ (converted to logit space as $-\log \tau$). Let $\mathcal{H}$ be the set of such hard pixels ($\tau$ is set to 0.7):

\begin{equation}
	\mathcal{H} = \left\{ i \mid \mathcal{L}_{\text{CE}}(i) > -\log \tau \right\}
\end{equation}

If $|\mathcal{H}| < N_{\text{min}}$, we select the top $N_{\text{min}}$ hardest pixels by loss. The final loss is the mean over these hard pixels:

\begin{equation}
	\mathcal{L}_{\text{OHEM}} = \frac{1}{|\mathcal{H}'|} \sum_{i \in \mathcal{H}'} \mathcal{L}_{\text{CE}}(i),
\end{equation}

where $\mathcal{H}' = \mathcal{H}$ if $|\mathcal{H}| \geq N_{\text{min}}$, otherwise it is set to the top $N_{\text{min}}$ pixels. Both the loss weights of the segmentation and PMM are set to 1.

\begin{algorithm}
\caption{PMM training after full-input warm-up}
\begin{algorithmic}[1]
\STATE Sample nonempty visible set $\mathcal{V}$ using the full/masked mixture above.
\STATE Compute student features with routing normalized over $\mathcal{V}$ and segmentation loss $\mathcal{L}_{\rm seg}$.
\STATE Set $\mathcal{L}_{\rm pmm}=0$.
\IF{$\mathcal{V}\ne\mathcal{M}$}
\STATE Sample $t\in\mathcal{M}\setminus\mathcal{V}$ uniformly.
\STATE Compute detached target $y_t$ with the full-context EMA teacher.
\STATE Predict $\hat z_t=g_\phi(Q_t,\hat{\mathbf{X}}_{\mathcal{V}})$ and compute cosine loss $\mathcal{L}_{\rm pmm}$.
\ENDIF
\STATE Update student and predictor using $\mathcal{L}_{\rm seg}+\mathcal{L}_{\rm pmm}$.
\STATE Update teacher: $\theta_\tau\leftarrow0.999\theta_\tau+0.001\theta_s$.
\end{algorithmic}
\end{algorithm}

\section{Implementation Details}
We train the proposed method on 8 A5000 GPUs for DELIVER, 4 for MUSES, and 4 for MCubeS. Initial learning rates are 3.5e$^{-4}$ (MCubeS), 4e$^{-4}$ (MUSES) and 4e$^{-4}$ (DELIVER), with a 10-epoch warm-up at 0.1×.
Training is conducted for 500, 200, and 200 epochs on MCubeS, DELIVER, and MUSES respectively.
Each run begins with a warm-up stage (200 epochs for MCubeS and 15 epochs for both DELIVER and MUSES), during which the model is trained using all modalities without any masking.
We use AdamW with batch sizes of 1 (DELIVER) and 4 (MUSES, MCubeS) per GPU. Input resolutions are 1024×1024 (DELIVER, MUSES) and 512×512 (MCubeS), with standard augmentations: random resizing (0.5–2.0), flipping, color jitter, blur, and cropping.
FuseAdapt uses rank 8 for DELIVER and MCubeS, and 4 for MUSES. DINOv2-B (for DELIVER and MCubeS)~\citep{oquab2023dinov2} and DINOv2-S (MUSES) serve as the frozen VFMs.
Standard LoRA with the low-rank matrices A and B such that A is initialized with a small random Gaussian (e.g., $\mathcal{N}(0, 0.01)$) and B is initialized as zero, ensuring the initial weight perturbation is zero.
The number of learnable tokens in the predictor is the same as the sequence length of the target feature.

\section{More Ablation Studies}

\paragraph{Rank $r$}
We evaluate the performance of the proposed method with different rank $r$. The experiments are conducted on MCubeS on the MCSS setting. As shown in Tab.~\ref{tab:ablate_r}, FuseAdapt significantly improves performance over a frozen VFM, with the best results obtained at rank 8.

\begin{table}[!ht]
	\caption{Results of different rank $r$ under MCSS setting on MCubeS. All models are trained with four modalities. }
		\centering
		\begin{tabular}{ccccc}
			\toprule
			$r$ & 0 & 4 & 8 & 16 \\
			\midrule
			RADN & 51.16 & 55.86 & \textbf{57.62} & 57.03 \\
			\bottomrule
		\end{tabular}
	\label{tab:ablate_r}
\end{table}

\paragraph{Number of Layers Equipped with FuseAdapt LoRAs}
We further evaluate the impact of the insertion frequency of FuseAdapt LoRAs in the VFM, where a module is inserted every $N$ layers. A larger $N$ indicates fewer inserted LoRAs. The experiments are conducted on MCubeS under the MCSS setting. As shown in Tab.~\ref{tab:ablate_num}, inserting more FuseAdapt LoRAs modules yields greater performance improvements.

\begin{table}[!ht]
	\caption{Results of inserting FuseAdapt LoRAs every $N$ layers under MCSS setting on MCubeS. All models are trained with four modalities. }
	\centering
	\begin{tabular}{ccccc}
		\toprule
		$N$ & 0 & 1 & 2 & 3 \\
		\midrule
		RADN & 51.16 & \textbf{57.62} & 56.38 & 55.87 \\
		\bottomrule
	\end{tabular}
	\label{tab:ablate_num}
\end{table}

\paragraph{Routing Fusion in FFNs}
We introduce independent, modality-specific routing modules at the FFN stage to enable controlled and dynamic cross-modal fusion.
Unlike attention layers that focus on intra-modal context, these routing blocks refine each modality’s representation by selectively integrating complementary signals from others, based on the current context.
This fusion occurs after independent encoding, preventing early entanglement and allowing the model to adaptively weight modalities depending on availability, especially beneficial under modality masking. Importantly, each modality has its own dedicated routing layer, which maintains structural disentanglement and minimizes interference from masked or undertrained modalities.

The routing module adopts a lightweight noisy gating mechanism, adapted from sparsely activated Mixture-of-Experts (MoE) designs.
Given hidden states, the router outputs modality-specific routing scores through a linear layer, and injects learned Gaussian noise via a second linear layer followed by softplus. The input size matches the feature dimension of the VFM output, while the output size corresponds to the number of training modalities. The resulting noisy logits enable soft, differentiable expert selection.
Routers are trained end-to-end as part of the main optimization, without any additional loss terms.

We evaluate the effectiveness of routing mechanisms. As shown in Table~\ref{tab:ablate_routing}, incorporating routing fusion into the FFNs consistently improves performance.
When no fusion is applied (``without routing''), the model processes each modality independently and therefore cannot benefit from cross-modal interactions.
Using a simple average fusion yields only marginal gains, as it assumes that all modalities contribute equally and ignores their heterogeneous characteristics.
In contrast, routing fusion adaptively selects and combines modality-dependent FFN experts, leading to a clear improvement (from 56.09 to 57.62 mIoU).
This demonstrates that learned routing effectively captures the distinct roles that different modalities play in multi-modal feature refinement.
We also observe that a shared router yields a competitive gain (57.24 mIoU), showing that routing is helpful even without modality-specific specialization.

Notably, unlike Mixture of Experts structures that perform token-level dynamic expert selection (e.g., top-K gating), our routing deterministically assigns each LoRA expert to a specific modality.
All available modality branches participate without top-$k$ selection or a load-balancing objective. Mixing weights remain token-dependent, and absent sources are masked before normalization.

\begin{table}[!ht]
	\caption{Results of various kinds of routing strategies under MCSS setting on MCubeS. All models are trained with four modalities. }
	\centering
    \begin{adjustbox}{width=0.5\columnwidth,center}
	\begin{tabular}{ccccc}
		\toprule
		   & wo. routing & average & shared router &w. routing \\
		\midrule
		RADN & 56.09 & 56.43 & 57.24 & \textbf{57.62} \\
		\bottomrule
	\end{tabular}
    \end{adjustbox}
	\label{tab:ablate_routing}
\end{table}

\paragraph{Additional VFMs}
To assess backbone sensitivity, we evaluate our method with two alternative VFMs, EVA-CLIP-B~\citep{sun2023eva} and NeCo-B~\citep{pariza2024neco}, on MCubeS (in MCSS setting with full-modality training). As shown in Table~\ref{tab:ablate_vfm}, the proposed approach maintains strong performance across all backbones:

\begin{table}[!ht]
	\caption{Results of applying various VFMs under MCSS setting on MCubeS. All models are trained with four modalities. }
	\centering
    \begin{adjustbox}{width=0.4\columnwidth,center}
	\begin{tabular}{cccc}
		\toprule
		   & EVA-CLIP & NeCo & DINOv2 \\
		\midrule
		RADN & 56.02 & 56.28 & \textbf{57.62} \\
		\bottomrule
	\end{tabular}
    \end{adjustbox}
	\label{tab:ablate_vfm}
\end{table}

\paragraph{Scalability of FuseAdapt}
We analyze the scalability of FuseAdapt under different backbones and input resolutions.

For DINOv2-B with the input size of 512$\times$512 and rank of 8, adding an additional modality increases the parameter count by 1.11M, which corresponds to 1.12\% of the full model.
The memory cost, including optimizer states under AdamW (in float32 type and without AMP), is 16.90 MB, and the computational cost is 4.8 GFLOPs, representing only 0.55\% of the total model FLOPs (873.96 GFLOPs).

For DINOv2-S with the input size of 1024$\times$1024 and rank of 4, each additional modality introduces 0.29M parameters (1.14\% of the full model), 4.50 MB of memory, and 18.12 GFLOPs, which amount to 2.01\% of the overall FLOPs (902.76 GFLOPs).

Although the cost of FuseAdapt grows linearly with the number of modalities, the relative overhead on parameters, memory, and computation remains very small.
These figures describe adaptation-module overhead, not the total cost of processing an additional sensor through the frozen encoder. End-to-end deployment measurements are reported in Appendix~\ref{app:efficiency}.

\end{document}